\documentclass[letterpaper, 10 pt, conference]{ieeeconf}  

\IEEEoverridecommandlockouts                              

\usepackage{formatting} 

\title{\LARGE \bf
DenseTact: Optical Tactile Sensor for Dense Shape Reconstruction 
}

\author{Won Kyung Do and Monroe Kennedy III 
\thanks{Authors are members of the ARMLab in the Mechanical Engineering Department, Stanford University, Stanford, CA 94305, USA. 
{\tt\small \{wkdo,monroek\}@stanford.edu.} 
The first author is supported by a fellowship from the Kwanjeong Educational Foundation. \edits{This work is supported by the National Science Foundation under Grant 2142773.} Youtube link for DenseTact: \url{https://youtu.be/nhQzhsjbcQA}}
}

\begin{document}

\maketitle
\thispagestyle{empty}
\pagestyle{empty}

\begin{abstract}

Increasing the performance of tactile sensing in robots enables versatile, in-hand manipulation. Vision-based tactile sensors have been widely used as rich tactile feedback has been shown to be correlated with increased performance in manipulation tasks. Existing tactile sensor solutions with high resolution have limitations that include low accuracy, expensive components, or lack of scalability. 
In this paper, an inexpensive, scalable, and compact tactile sensor with high-resolution surface deformation modeling for surface reconstruction of the 3D sensor surface is \edits{presented}. \edits{By observing the contact surface with a fisheye camera, it is shown that the surface deformation can be estimated in real-time (18 ms) using deep convolutional neural networks.} This sensor in its design and sensing abilities represents a significant step toward better object in-hand localization, classification, and surface estimation all enabled by \edits{calibrated,} high-resolution shape reconstruction.
\end{abstract}


\section{INTRODUCTION}


Robots capable of manipulation have been adopted in a variety of industrial applications, including car manufacturing, welding, and assembly line manipulation. Despite the surge of robotics in automation, it is still challenging for robots to accomplish general, dexterous manipulation tasks. Increased robotic dexterity particularly of small objects, would impact across society from agility of manufacturing to collaborative robots for assisted living \cite{smith2007caring, Ansah2014}. 
Modern robots capable of performing manipulation still find it challenging to perform in-hand manipulation or accurately measure the deformation of soft fingertips during complex dexterous manipulation. This problem has been managed from many different approaches, including designing new hardware, developing perception and recognition functionality for the robots, or solving control and motion planning tasks of the robotic assistant. However, persistent challenges for manipulation in general tasks are the lack of tactile feedback with sufficiently high resolution, accuracy, and dexterity that rivals anthropomorphic performance.
\begin{figure}[t!]
	\centering
	\includegraphics[width=2.7in]{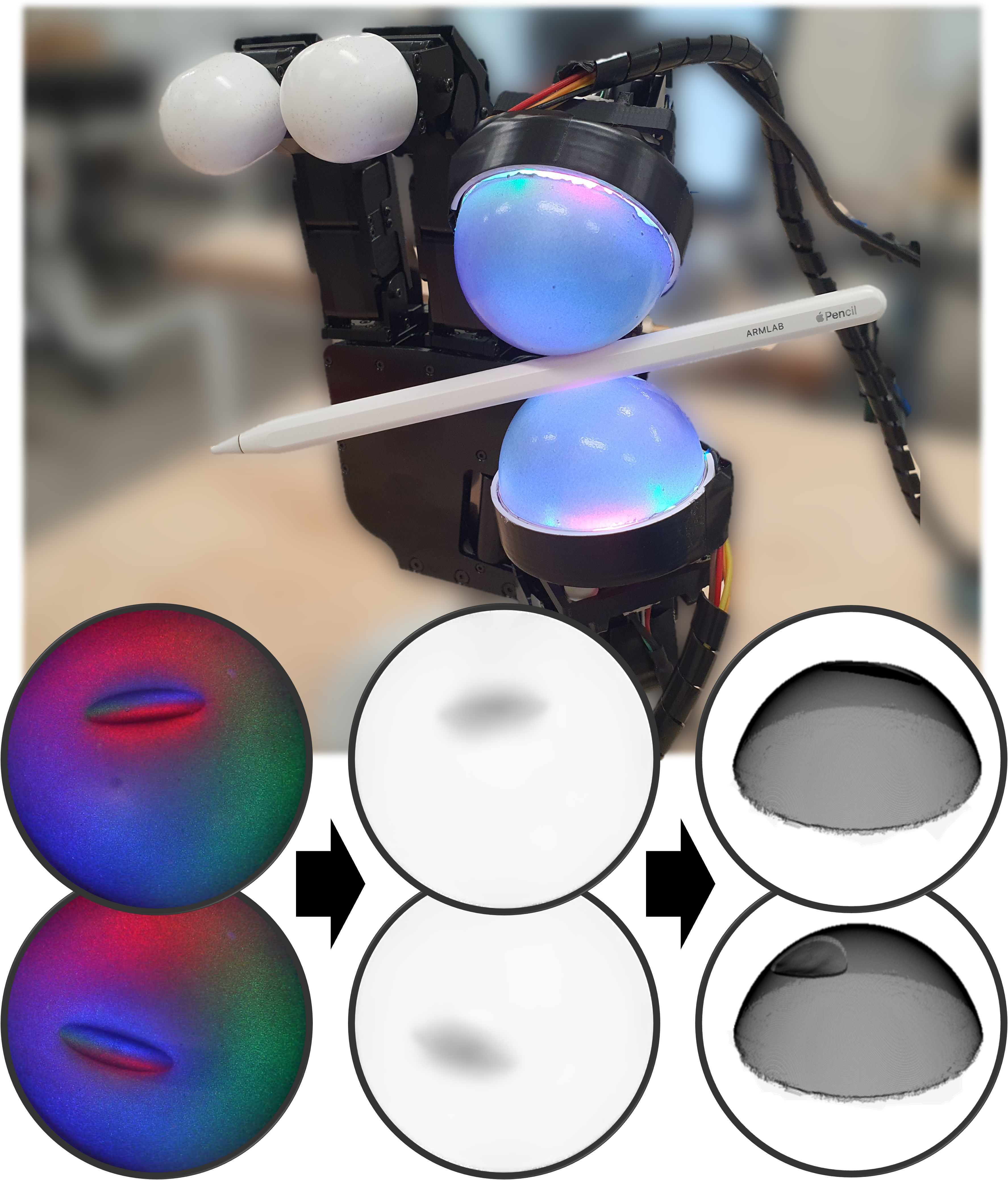}
  \caption{\textbf{DenseTact.} Sensor is outfitted on the Allegro\texttrademark{} hand. Left bottom image shows the image taken from the cameras. Middle bottom shows the resultant depth image. Right bottom represents the 3D reconstructed surface of current sensor.}
  	\label{fig:main}
      \vspace{-0.67cm}
\end{figure}
Vision-based, soft tactile sensors capable of reconstructing surface deformation with high accuracy can address some of the above problems. To better model the object(s) within the manipulation task, precise geometric and contact information of the object and environment must be ensured. Therefore, high-resolution tactile sensing that provides rich contact information is an essential step towards high precision perception and control in dexterous manipulation.

Common major challenges with the vision-based tactile sensor are accurate pose estimation from sensory input, limited 2D-shape of the sensor, and high construction costs. Existing tactile sensors such as Gelslim and Gelsight can detect the surface deformation from the soft elastomer, but the 2D shape limits in-hand manipulation. 
This work presents a tactile sensor that has a 3D shape, is relatively inexpensive, and is capable of solving state estimation tasks more accurately. The developed sensor is a vision-based, soft tactile sensor designed for accurate position sensing.
The developed sensor design has a hemispherical shape with an inexpensive, fisheye lens camera with a soft elastomer contact surface (Fig. \ref{fig:main}). The interior of the sensor is illuminated, allowing for tactile feedback with a single image. From the high-resolution image, the corresponding 3D surface of the sensor is estimated. Practical applications of the sensor include in-hand pose estimation of a held object. 


\edits{
Beyond the 3D-shape sensor design with low-construction cost and high sensing resolution, our major contributions include: 
\begin{inparaenum}[1)]
    \item \textbf{novel and dense calibration process for model which performs 3D shape reconstruction from single image with high resolution}, and   
    \item \textbf{adapted deep neural network architecture for estimating sensor surface from interior sensor image.}  
\end{inparaenum}
}

The rest of this paper is organized as follows: related work is discussed in Sec. \ref{sec:related_works}, Sec. \ref{sec:sensor_design} presents the sensor design, Sec. \ref{sec:shape_reconstruction} presents the method of shape reconstruction calibration and modeling, Sec. \ref{sec:results} shows shape reconstruction model results, and the conclusion and future work is discussed in Sec. \ref{sec:conclusion}. 

  \begin{figure}[t]
  
    \vspace{0.2cm}
      \centering
      \includegraphics[width = 2.75 in]{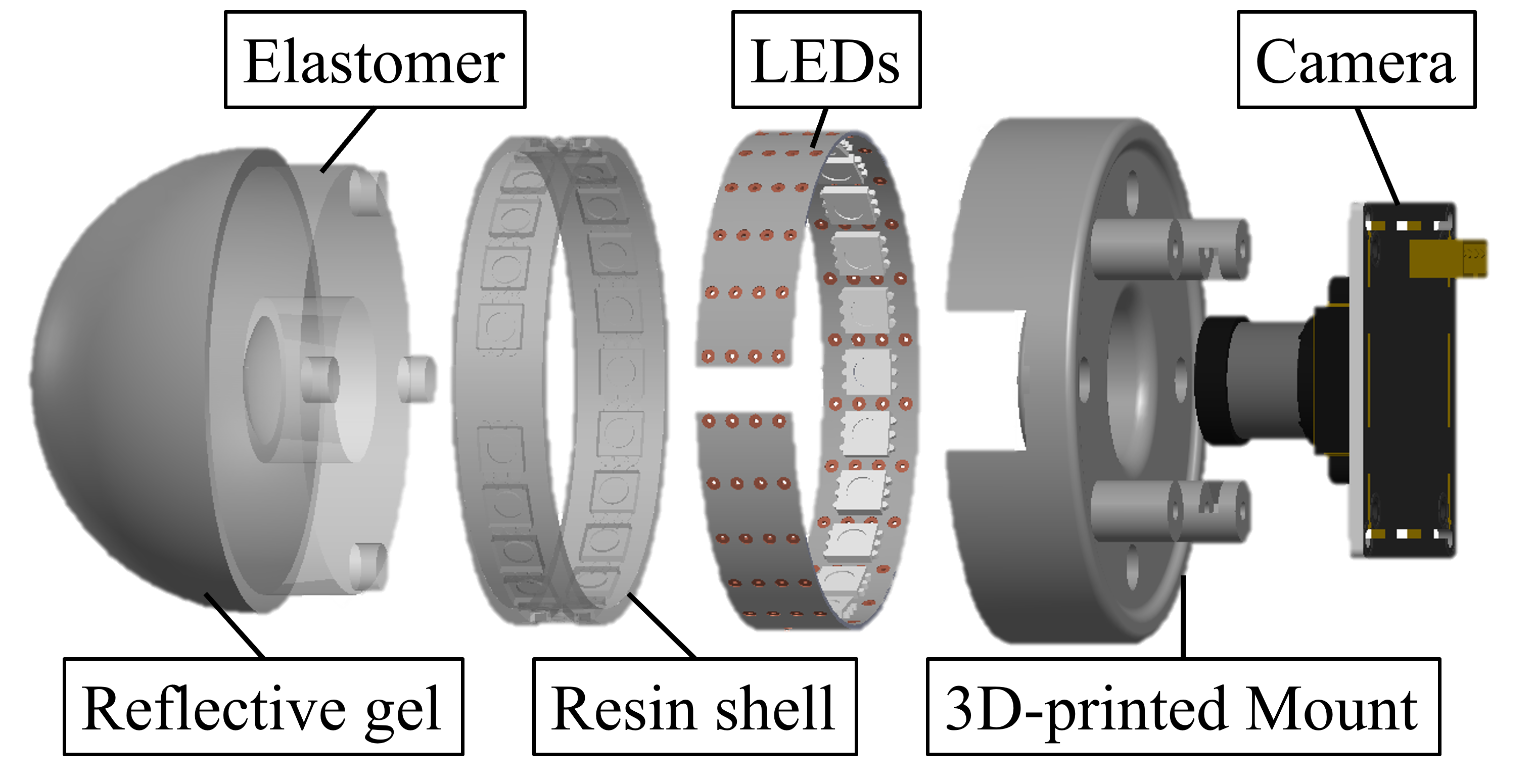}
      \caption{ \textbf{DenseTact Design}. Exploded view of DenseTact sensor. The soft sensor is composed of a silicone elastomer coated with a reflective surface and attached on a 3d-printed-mount with a fisheye lens camera and LED strip. }
      \label{fig:CAD}
      \vspace{-0.6cm}

  \end{figure}



\section{Related Works}
\label{sec:related_works}


The ability for robots to perform complex dexterous manipulation is inherently dependent on the quality of the tactile sensors. The tactile sensor's resolution, noise, bias, and repeatability all contribute directly to the robot's ability to perform robust object manipulation and control. A variety of tactile sensors have been developed with many different approaches including piezoelectric \cite{wettels2009multi}, optics \cite{Dong2017}, resistance \cite{cheng2009flexible}, or capacity \cite{Huh2020}. 
Recently, vision-based tactile sensors have been become very popular compared to previous sensors due to their higher resolution. Optical sensors have proven valuable for tactile sensing with usefulness in detecting both shear and normal forces \cite{Ma2019}, controlling grasping motion \cite{doi:10.1177/02783649211027233}, or adjusting grasp \cite{Hogan2018}. Table \ref{table:comparison} compares vision-based sensor design shape, resolution, and calibration methodology with the proposed sensor of this work in the final row.

However, most vision-based tactile sensors are expensive, bulky, and limited to 2D shapes. Tactile sensors such as Gelsight, Gelslim, or DIGIT have high resolution, but a flat surface limits the application in a manipulation task \cite{Dong2017, donlon2018gelslim, lambeta2020digit}. The sensors with air-pressure-based approaches provide the elastic, restorative reconstruction of surface during contact. However, sensors have large sizes for pneumatic transmission \cite{mcinroe2018towards, Alspach2019}.
Leveraging surface normal from image intensity is a common method to perform depth reconstruction in a limited environment \cite{hogan2021seeing, johnson2011microgeometry, woodham1980photometric, Dong2017, ferrier2000reconstructing }. Assuming that the surface is Lambertian, the linear reflective function between the pixel's intensity with the surface normal can be obtained, making the lookup table to get the surface normal. However, the limitations on dimensions of the lookup table and assumption of Lambertian surface make the approach inapplicable for the 3D-shaped sensor.

The tactile sensors with 3D curved surface designs such as Omnitact\cite{padmanabha2020omnitact} allow multi-directional sensing but are costly, especially for multi-finger applications. The round fingertip sensor from \cite{romero2020soft} also has a 3D shape but relatively small resolution compared to other vision-based sensors. \edits{To solve the current issue from previous sensors, the design in this work is a cost-efficient sensor} with a 3D shape that enables high-resolution sensing feedback towards robot manipulation.




\begin{table}[t]\centering
    \vspace{0.2cm}

\begin{tabular}{| c || c | c | c | c|}
 \hline
Name&3D shape&Resolution&Calibration\\ \hline
 \begin{tabular}{@{}c@{}}Gelsight-like \\ sensors \cite{s17122762, taylor2021gelslim3, patel2020digger} \end{tabular}  &$\times$& $640 \times 480$ & 3D \\\hline
 Softbubble\cite{Alspach2019} & \checkmark &$224 \times 171$   & 3D \\\hline
 Omnitact\cite{Padmanabha2020}& \checkmark   & $400 \times 400$& $\times$ \\ \hline
 NeuTouch\cite{Taunyazov2020}& $\times$ &  $39$ & $\times$ \\ \hline
 TacTip\cite{ward2018tactip}&  \checkmark  &  $127 ~ 180$  & $\times$ \\ \hline
 Optofiber-sensor\cite{Baimukashev2020}&$\times$ & $61$ fibers & 2D   \\ \hline
 Digit\cite{lambeta2020digit}& $\times$  &  $640 \times 480$ & $\times$ \\ \hline
 FingerVision \cite{9420089} & $\times$  &  $640 \times 480$ &  3D\\ \hline
 GelTip \cite{gomes2020geltip} & \checkmark & not specified &  2D\\ \hline
 \textbf{This work: DenseTact} & \checkmark & $800$ $\times$ $600$ & 3D \\
 \hline

\end{tabular}
\caption{\textbf{Related Work.} List of high-resolution vision-based tactile sensor for shape reconstruction\edits{. 2D-calibrated sensors localize the surface position without estimating depth.}}
\label{table:comparison}
     \vspace{-0.7cm}

\end{table}

\section{DenseTact Sensor Design}
\label{sec:sensor_design}
\subsection{Design Criteria}

For a general manipulation task, an accurate tactile sensor can maximize the quality of information at the contact surface. To meet this criterion, the design goals for the sensor are:
\begin{inparaenum}[1)]
    \item Small sensor size, useful for in-hand, small-object manipulation.
    \item 3D curved shape with a very soft surface, enabling versatile manipulation. 
    \item High-resolution surface deformation modeling (shape reconstruction) for contact sensing. 
\end{inparaenum} A fisheye-lens camera combined with hemispherical shaped, transparent elastomer design satisfies the above criteria. When the contact boundary of the elastomer has a reflective coating, the monocular camera can observe the internal deformation of the elastomer from a single image provided there is sufficient interior illumination.


\subsection{Fabrication}\label{sec:fab}

\subsubsection{Elastomer Fabrication}
Primary design goals for the vision-based sensor are that it be cost-efficient, high-resolution, and has a 3D shape which is captured by a hemispherical design useful for soft-finger manipulation. This motivates the design selection of a hemispherical shaped, transparent elastomer with a reflective surface boundary to allow an interior monocular camera to observe the sensor deformation.
 The elastomer selected is the extra-soft silicone for the elastomer (Silicone Inc. P-565 Platinum Clear Silicone, 20:1 ratio). It has a 6.5 shore A hardness, similar to the hardness of human skin. The extra soft elastomer maximizes the surface deformation even from a small shear force. A clear elastomer surface is ensured using the marble to make a silicone mold of the elastomer. After the elastomer is cured, Inhibit X\texttrademark{} is applied \edits{as an adhesive} before airbrushing the surface with the mixture of reflective metallic ink and silicone (Smooth-on Psycho Paint\texttrademark{}).

\subsubsection{Camera and Illumination system}

A cost-effective camera solution with a Sony IMX179 image sensor (8MP, 30fps) is selected with a small size for in-hand manipulation while enabling real-time image processing. We select a circular fisheye lens with 185$^\circ$ FoV to cover the whole hemispherical-shaped elastomer. 
The interior of the soft-sensor is illuminated by an LED strip with a flexible PCB. The LED strip contains 24 RGB LEDs arranged into a cylindrical pattern at the base of the elastomer. 
\edits{Similar to work done in \cite{s17122762}, we activate 3 LEDs from LED strip.} This illumination strategy allows surface depressions to emit color patterns that indicate surface shape correlated to color channel reflectivity.


\subsubsection{Sensor Assembly}
\figref{fig:CAD} and center image in \figref{fig:raycast} show the exploded and cross-sectional view of the sensor respectively. Once assembled, the center of the camera lens is aligned to coincide with the center of the hemispherical elastomer, and the strip LEDs are located right below the elastomer as shown in \figref{fig:CAD}. 
DenseTact has a height of 35mm, and the hemisphere elastomer has a radius of 25mm. In this configuration, DenseTact can sense almost 4,000mm$^2$ of the elastomer surface. A 3D printed camera mount is used to fasten the camera and elastomer, and the LED strip is fitted around the interior of this camera mount.
A major problem of elastomer-based tactile sensors is their durability. Resin is used as an adhesive for the LEDs and the camera mount to increase durability. Once the resin is cured, the cured elastomer is attached to the mount using Sil-Poxy adhesive. This construction method allows the elastomer to support higher forces without fatigue and degradation.
Preliminary experiments have shown that this DenseTact sensor design can be reduced in size, with a hemispherical radius of 15mm and 25mm sensor height, enabling more delicate manipulation.


The cost for the entire sensor is less than \$80, and most of the sensor cost comes from the camera system (\$70). LED strip (\$5.5), elastomer with reflective surface including resin part (\$ 3.5), and camera mount (\$1) cost \edits{in} total \$10. Furthermore, the whole fabrication procedure only takes less than two days (mostly come from elastomer curing), which enables the fast application of the robot manipulation. 

\section{Shape Reconstruction}
\label{sec:shape_reconstruction}
Shape reconstruction from sensor interior surface normal estimation has been used for Gelsight-similar, vision-based sensors \cite{johnson2011microgeometry, Yuan2017, romero2020soft, taylor2021gelslim3}. Assuming the sensor surface is Lambertian and the reflection and lighting are equally distributed over the sensor surface, the intensity on each color channel can be leveraged to estimate a unique surface normal. The intensity color difference between the deflected and non-deflected surface is then correlated to the surface gradient $ \nabla f(u,v)$ by inverting the reflective function R. 
\textit{Gelsight-like sensors calibrate the $R^{-1}$ by mapping RGB values to the surface normal through a lookup table.} From the surface normal for each point, the height-map of the sensor can be reconstructed by solving the Poisson equation with the appropriate boundary condition. 

However, \edits{the} above method is not applicable for 3D shape sensor with non-Lambertian surface. Then the reflective function R also depends on the location as well as surface normal:
\begin{equation} I(u,v) = R \left( \frac{\partial f}{\partial u}(u,v),\frac{\partial f}{\partial v}(u,v), u,v \right), \label{eqn:intensity} \end{equation}
where $(u,v)$ is the position of the pixel or corresponding location on the sensor surface. The above function is nonlinear and it is difficult to get the inverse. One way to obtain the inverse function is to leverage a data-driven approach. Given that the sensor environment is restricted and reproducible, if the sensor surface shape is known (ground truth) then to perform shape reconstruction the objective is to determine a nonlinear function M such that
\begin{equation}
   (R, \theta, \psi) = M(I_{rgb}(u,v)), 
   \label{eqn:map_intensity_inverse_M}
\end{equation}
where $ (R, \theta, \psi) $ is corresponding spherical coordinate of the sensor surface from the position in the image frame $(u,v)$. \textit{One way to solve equation \ref{eqn:map_intensity_inverse_M} \edits{is to model $M$ with a representative network.}}

\subsection{Depth Data Generation}
  \begin{figure}[t!]
      \vspace{0.2cm}

      \centering
      \includegraphics[width = 2.8in] {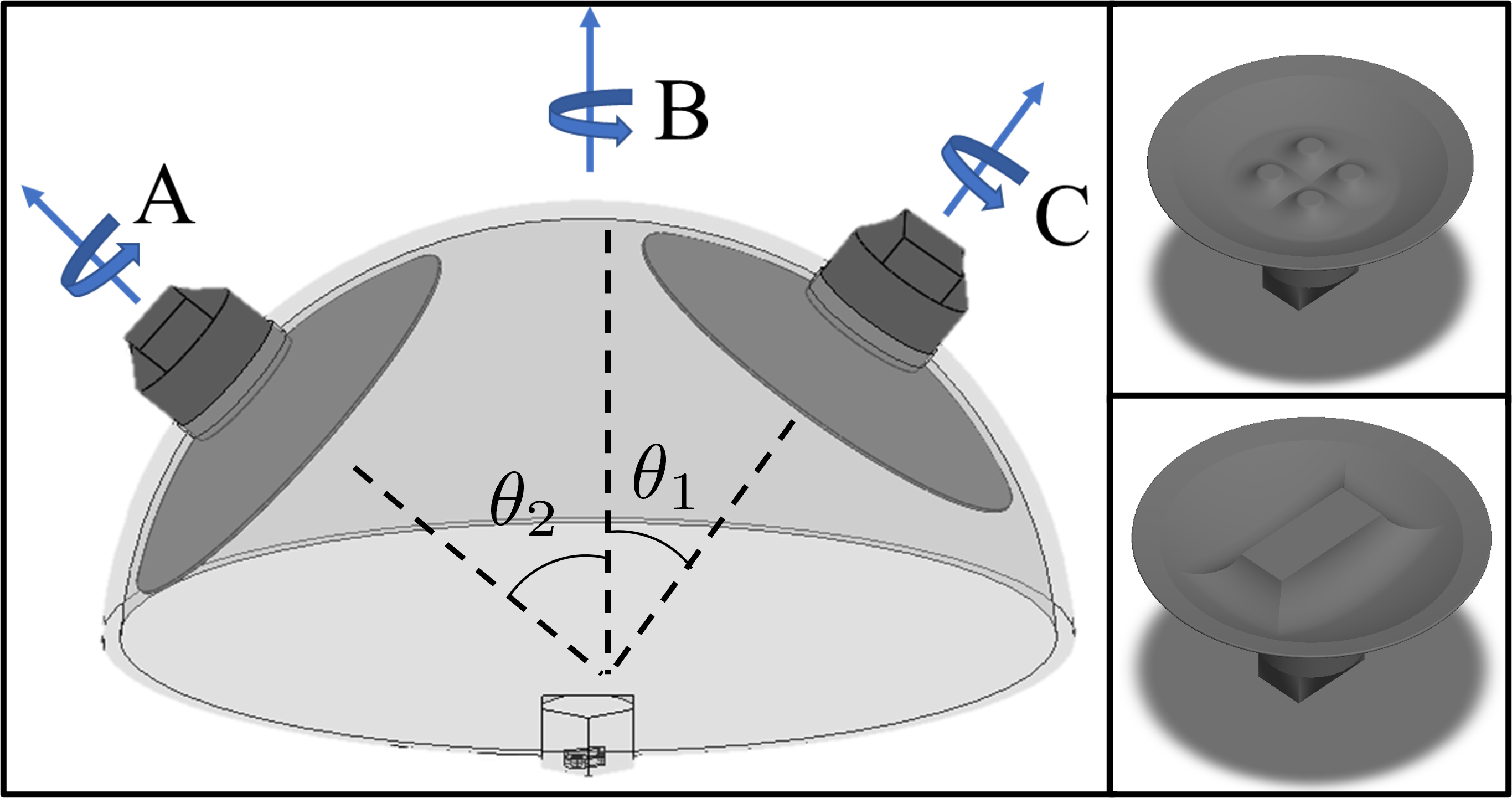}
      \caption{\textbf{Data Collection.} 3D printed parts are printed and pressed in varying configurations into the sensor for calibration.}
      \label{fig:datacollect}
          \vspace{-0.7cm}

  \end{figure}
  
 DenseTact must produce a high-resolution representation of the sensor surface from a single image.
This requires accurate, high-resolution ground-truth surface knowledge for model training. The sensor surface is hard to externally detect using commercial range-finding, active sensors (projected light or time-of-flight sensors). Measurements from such range-finding sensors have errors at millimeter scale, which would propagate to the ground-truth measurement.
To avoid this problem, data is generated by 3D printing a known computer-generated surface model and is subject to the accuracy of the 3D printer. The dataset generated in this manner has all the shape information required to estimate the surface shape at each corresponding image pixel necessary for model training.
Given that partial contact would cause deformation of the soft-sensor in un-touched regions of the elastomer, the known 3D printed shapes pressed into the elastomer cover the entire sensor surface at once, limiting unwanted motion, such that every location on the surface is known for a given contact. 
Training shapes were printed on \edits{an} Ultimaker S5 3D printer and included an indicator (large shell shown in \figref{fig:datacollect}) as well as indenters (small components in \figref{fig:datacollect}). \edits{A full set of indenters and indicators selected can be found on \footnote{\url{https://github.com/armlabstanford/DenseTact}}github.} 
The Ultimaker\texttrademark{} S5 is capable of printing layers with a minimal size of 60 microns and positioning precision of 6.9 microns.
The stated uncertainty of the fused deposition modeling (FDM) printer is 100 microns (0.1mm), however this still presents a valid method of obtaining a shape ground-truth for the entire surface at once under depression.
To diversify the generated dataset from a limited source of material, a hemispherical shape indicator is printed with a single or multiple holes and indenter that fits the hole. The right two images in \figref{fig:datacollect} show examples of the indenter. A total of \edits{37} different hemispherical indicators were printed with a different number of holes and locations. Indicators have different shapes variations that include a single or two holes or with a hole and in-built indenter. The left image in \figref{fig:datacollect} shows the overall view of the printed indicator that has two holes assembled with two different indenters. Location of hole varies with different $\theta_1$ and $\theta_2$. A total of \edits{25} different indenters with various shapes were also printed and examples are shown in the right two images of \figref{fig:datacollect}. The dataset is also varied by changing the angle in terms of \edits{axis A, B, and C} in \figref{fig:datacollect}. \edits{We automated the rotation on the B axis with stepper motor (400 step/rev) during data collection with CNC machine, but rotation on A and C axis are manually adjusted. Total data collection takes ~40 hours for this generation. We rotated the indicator in 0.9$^\circ$ per each data on axis B while rotating 45$^\circ$ on axis A and C.} 

For accurate position control of 3D printed objects, a CNC machine is utilized for data generation. The sensor is fixed on the bottom plate while the 3D printed object is mounted on the motorized part. \edits{CNC machine moved in z-direction pressing into the sensor.} 
After getting the radial value from the STL file with the ray casting algorithm, the depth value is normalized to an 8-bit integer (0-255) to reduce the size of the output value. Possible resolution loss is prevented by normalizing the value with the max depth depression (9.4mm). By doing so, 1-pixel intensity corresponds to the 0.0354mm increment in actual depth value.
Finally, a total of 30,200 different contact configurations are generated which consists of 29,200 training configurations and 1,000 test configurations. Note that a different combination of indicators and indenters is used for the test set comparing to the training set. The total dataset size is 3.6 GB.

\subsection{3D correspondence from camera image}
    \begin{figure}[tb]
        \vspace{0.2cm}

      \centering
      \includegraphics[width =3.1in]{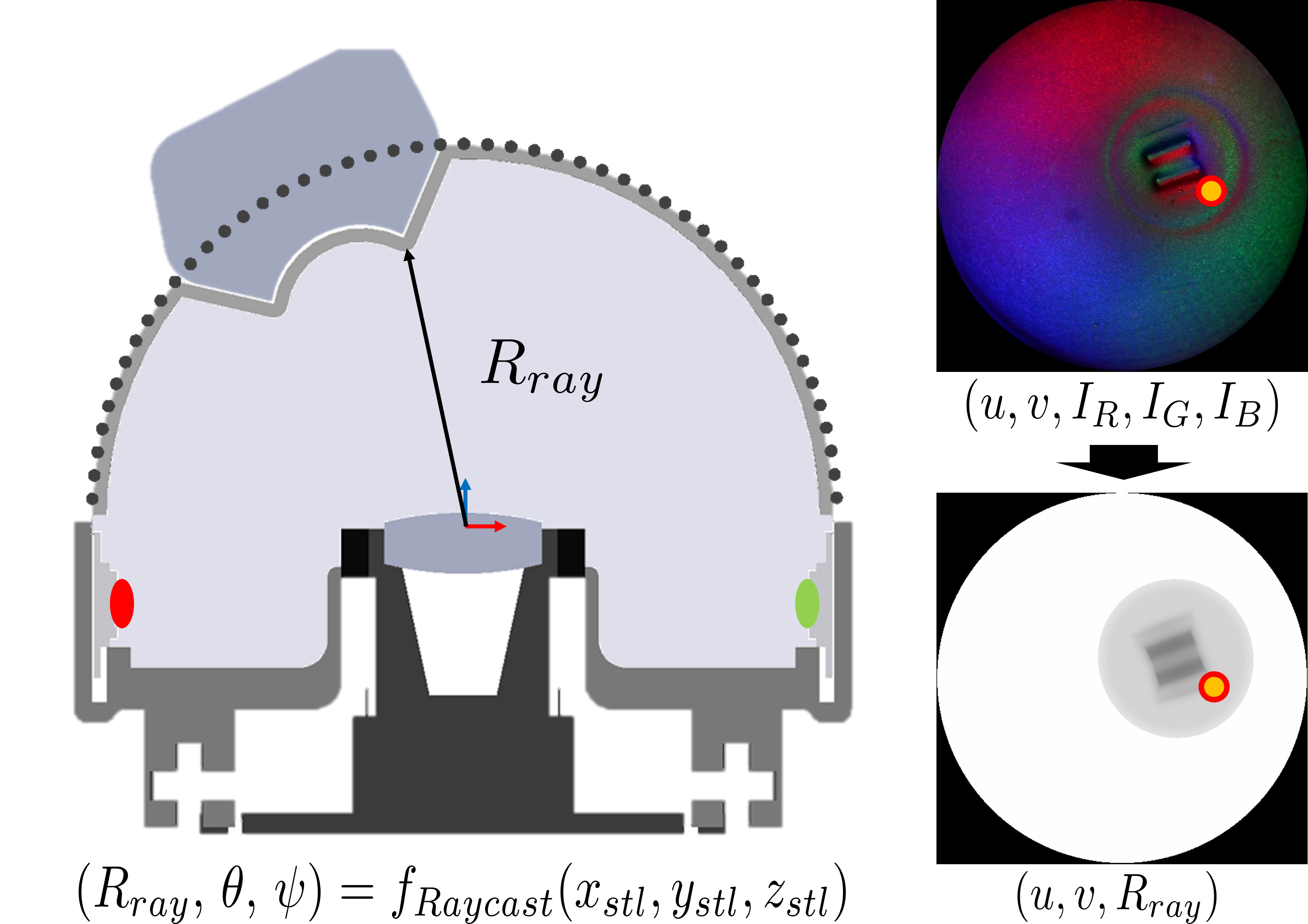}
      \caption{\textbf{Ray Casting Algorithm}. Ray casting is used to determine the radial depth from the 3D calibration surface which is then projected into the image plane.}
      \label{fig:raycast}
          \vspace{-0.6cm}

  \end{figure}

 \begin{figure*}[t]
      \centering
          \vspace{0.2cm}
    \sbox\twosubbox{%
      \resizebox{\dimexpr.97\textwidth-1em}{!}{%
        \includegraphics[height=2.6in]{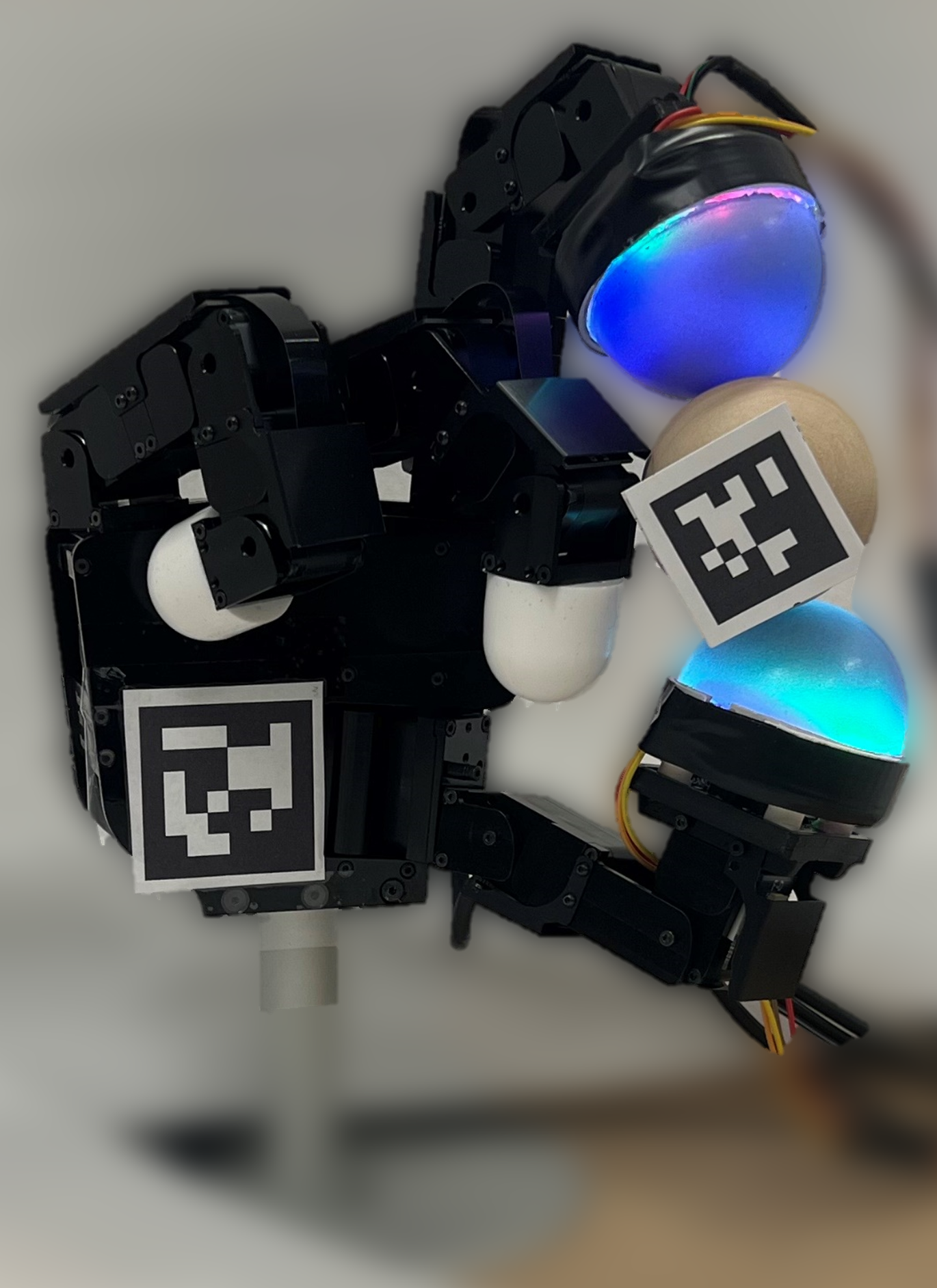}%
        \quad
        \quad
        \includegraphics[height=2.6in]{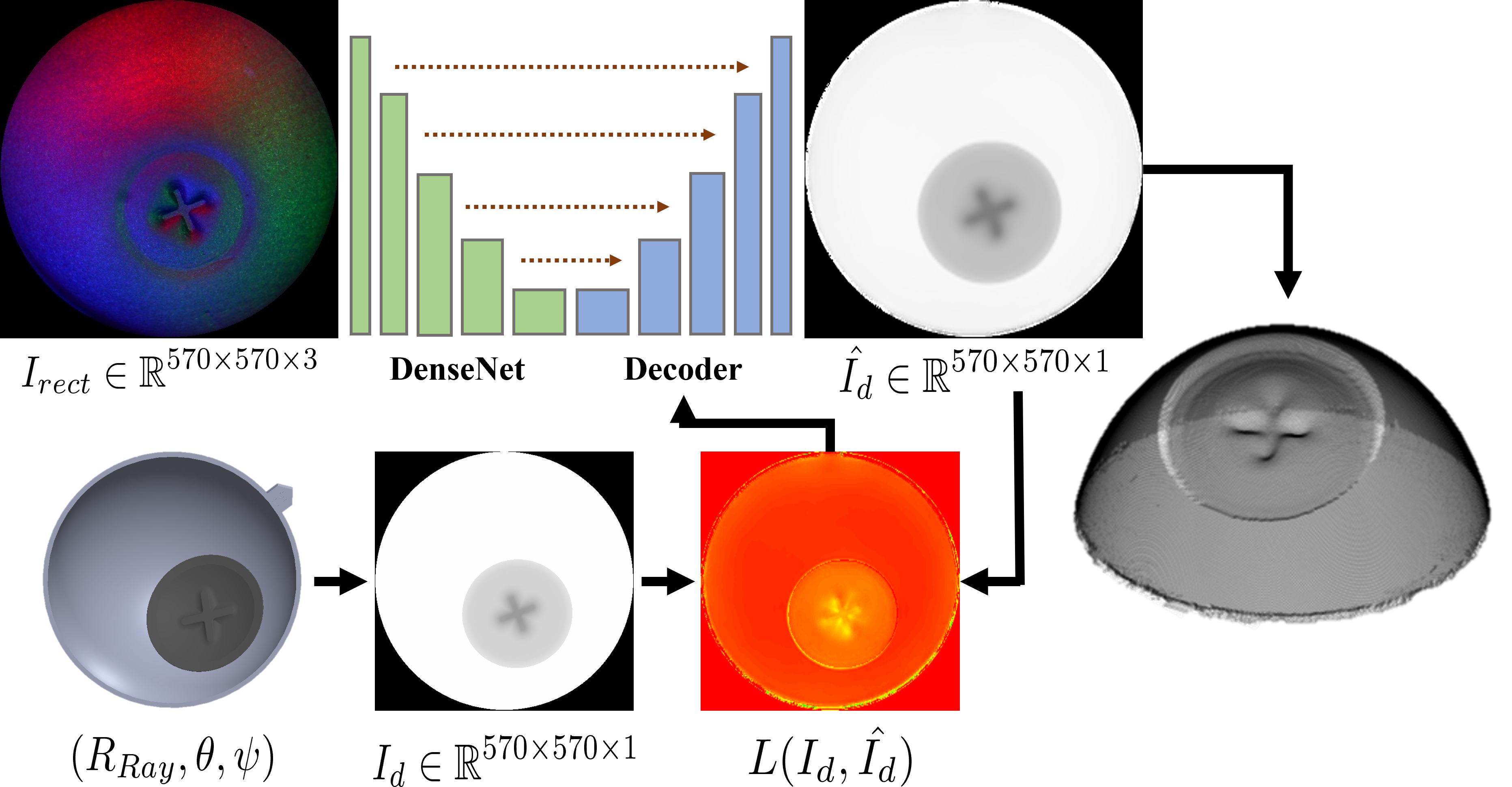}%
        }%
    }
    \setlength{\twosubht}{\ht\twosubbox}
    \subcaptionbox{ DenseTact with Allegro hand \label{fig:process_a}}{%
      \includegraphics[height=\twosubht]{figures/procedurefig_a.png}%
    }\quad
    \subcaptionbox{Network Architecture \label{fig:process_b}}{%
      \includegraphics[height=\twosubht]{figures/procedurefig_b_v2.png}%
    }
      \caption{\textbf{ DenseTact Algorithm}. The sensor interior is the input to the autoencoder network. The ground truth is provided from the object CAD model and converted to a depth image. The resultant disparity between prediction and ground truth is used to train the network. The output depth image is converted to a 3D point-cloud via a correspondence step.}
      \label{fig:process}
          \vspace{-0.45cm}

  \end{figure*}

The next step for shape reconstruction is finding correspondence between the image from the camera and sensor surface. The fisheye lens produces additional distortion on the image and bars the use of common calibration methods since correspondence with the 3D-shaped sensor surface is needed. The calibration method for a wide fisheye lens and the omnidirectional camera has been proposed in \cite{scaramuzza2006toolbox}, however, the main purpose of the calibration is to get the undistorted panoramic image. Therefore, a new correspondence model for the sensor must be \edits{developed}. 

First, a 3D-printed indicator with a known size is built. The indicator has 2mm thickness and a saw-tooth shape with equal angular interval, $5^{\circ}$. By simply pushing the indicator in a fixed position parallel with the x-axis of the image, the saw tooth is detected from the sensor image. The position of each saw-tooth from the image is detected by using the Canny edge method. From these detected edges from the image, the edge position is matched with the edge position on the sensor surface. 

The distorted image from the camera has symmetric distortion in terms of $\psi$ direction, and the center of the image is aligned with the center axis of the tactile sensor. The radius from the center of the image corresponds with the $R\,sin(\theta)$ in the hemispherical sensor surface. The Gaussian Process (GP) regression model is implemented for the correspondence between radius in image $r$ and radius in the sensor surface $R\,sin(\theta)$. From this correspondence, these indexes are matched and each image pixel is transformed into the right $\theta, \psi$ in spherical coordinates. 
\begin{equation} R\,\,sin(\theta) = f_{GP}(r(u,v))\label{eqn:r_sin_theta}\end{equation}
\begin{equation}
(\theta, \phi) = \left(sin^{-1} \left(\frac{f_{GP}(r)}{R} \right), tan^{-1}\left(\frac{v-v_c}{u-u_c} \right) \right) \label{eqn:sphere_uv_map_gp} \end{equation}
where $u_c,v_c$ is center of the image plane.
Once the conversion between $(u,v)$ and $(\theta, \phi)$ is done, the corresponding R from the STL file of a combined 3D indicator is found. Based on the vector generated from $(\theta, \phi)$ for each pixel, the ray casting algorithm allows computing the closest point on the surface of triangular mesh from STL file \cite{Zhou2018}. Fig. \ref{fig:raycast} shows the cross-section view of the sensor with a radius defined from the ray casting algorithm. Once the depth or the radius information for each corresponding pixel value is obtained, the pixel value is directly matched with the depth from the ray casting algorithm:
\begin{equation}R_{ray}(u,v) = f_{raycast}(Mesh_{stl}, \theta(u,v), \psi(u,v)), \label{eqn:raycast} \end{equation}
which makes 1:1 correspondence between input images. Each image from the sensor has $800\times 600$ pixel resolution. First, the image is cropped and the useful pixel value is extracted from the above GP and ray casting algorithm. A total of 253,213 useful pixel values are extracted from the single image and the ground truth image is reconstructed (right images of Fig. \ref{fig:raycast}). 

\subsection{Modeling}




  \begin{figure}[t!]
      \centering
          \vspace{-0.1cm}

      \includegraphics[width = 3.1 in ] {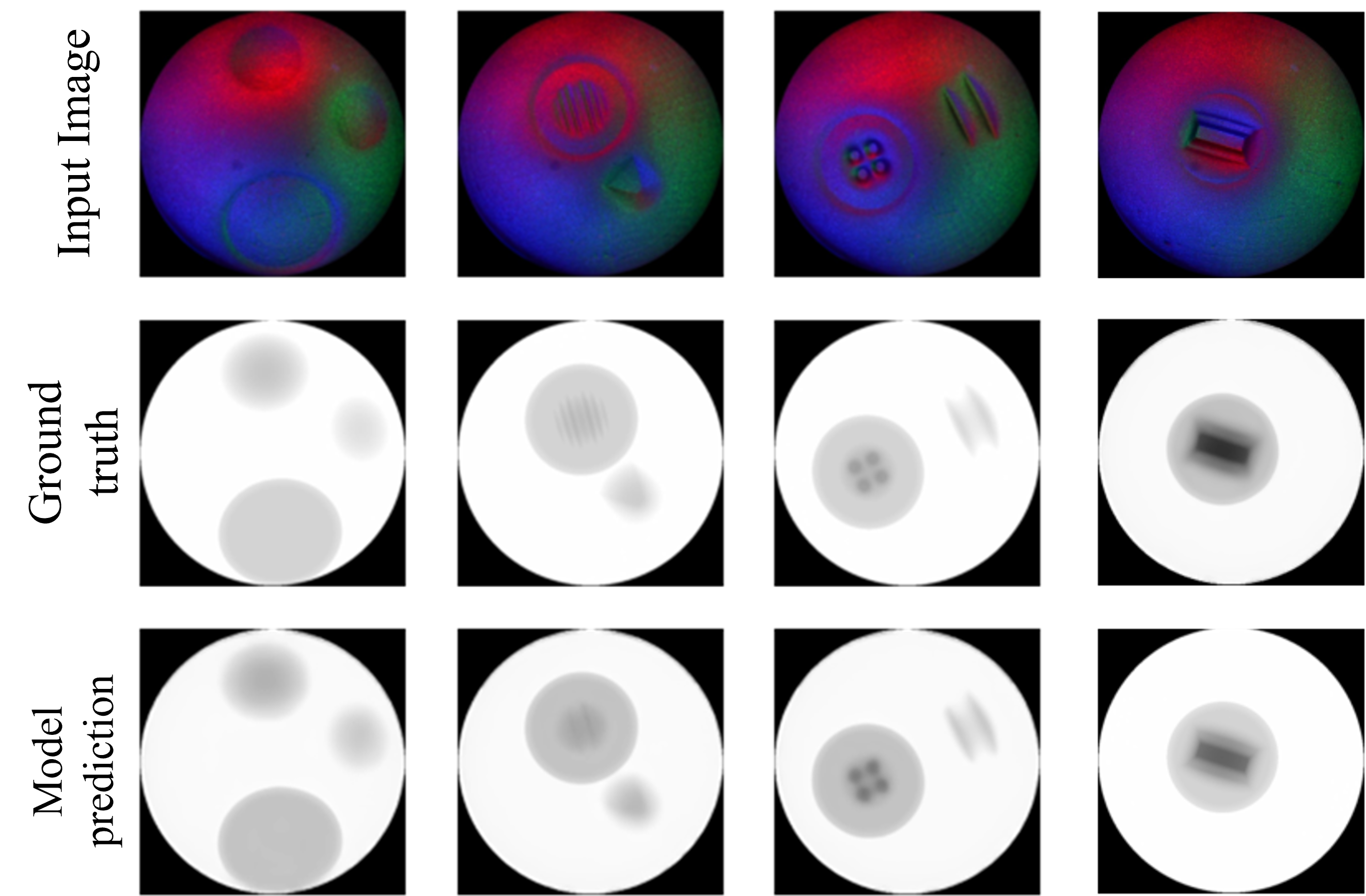}
      \caption{ \textbf{Shape Reconstruction Performance.} Select examples from test set for sensor shape reconstruction.}
      \label{fig:dataresult}
          \vspace{-0.6cm}

  \end{figure}

The goal of the model is to estimate the depth image from the RGB input image with the same dimension. This can be interpreted as the single image depth estimation problem, but unlike similar implementations, contextual semantic knowledge from the input image is unavailable in this case. Some of the leading network strategies are leveraging encoder-decoder structure with additional network block that utilizes global information from the depth image \cite{bhat2021adabins}. Unlike general depth images from datasets such as \cite{Silberman:ECCV12}, the DenseTact dataset requires more focus on the local deformation information since the global information is similar between datasets. Therefore, a much simpler version of the network is selected with a pre-trained encoder and decoder with skip connections \cite{alhashim2018high}. 
The encoder part of the network consists of a pre-trained DenseNet-161 \cite{huang2017densely}. The decoder part concatenates the previous up-sampled block with the block of the same size from the encoder, which allows learning local information by skipping the network connection. The implemented loss is the combination of three losses: point-wise L1 loss on depth values, L1 loss on the gradient of the depth image, and structural similarity loss as in \cite{wang2004image}. 
Fig. \ref{fig:process} shows the overall working process of the sensor with network structure. The image size has been resized from $570\times 570 \times 3$ to $640 \times 480 \times 3$ before it pass the network and resize the output result $320\times 240$ to $570\times 570 $. Furthermore, the training quality is maximized by re-scaling the range$(0,255)$ depth value into range$(10,1000)$. The network is trained with 16 epochs (460K iterations) with batch size 4 from NVIDIA P100 16GB GPU. 

\section{Results and Discussion}
\label{sec:results}
\subsection{Results}



\begin{figure}[t!]
      \centering
                \vspace{0.2cm}

      \includegraphics[width = 2.9 in] {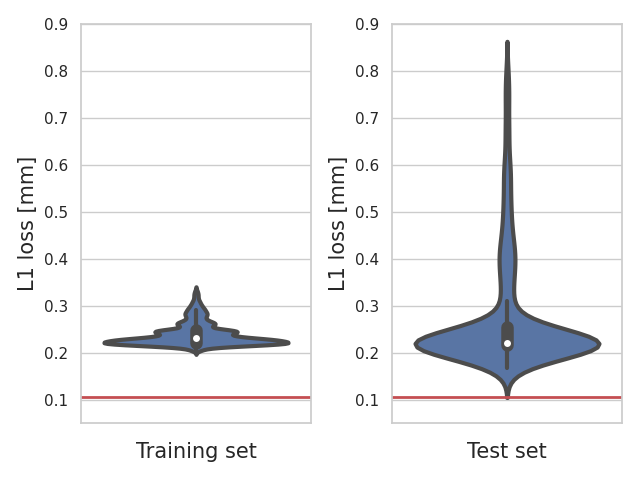}
      \caption{\textbf{Re-projection error}. Each data point represents the mean $L1$ re-projection error for a single image (effectively 253,213 pixels). Statistics are shown for the training (29,200 images) and test sets (1000 images).}
      \label{fig:l1result}
           \vspace{-0.65cm}
  \end{figure}

Fig. \ref{fig:dataresult} shows the qualitative result of shape reconstruction. The first row is the input image and the second row is the ground truth from the depth image. The model prediction is shown in the third row. Results show that the reconstruction works fairly well with the given single monocular image. Note that input images come from the test set image, so the model did not train the input image. Furthermore, the sensor only takes 18.17 milliseconds on average to predict the depth view from a single image. This implies that the sensor is able to perform real-time manipulation tasks with 30fps. The point cloud is reconstructed based on the registered index with a predicted depth value. The right image in Fig. \ref{fig:process} and right-bottom images in Fig. \ref{fig:main}shows the reconstructed sensor surface from depth image. After reconstruction, the point-wise L1 loss on the predicted depth image between the test set and training set are compared. Fig. \ref{fig:l1result} shows the violin plot of the reconstruction error statistics of all training and test sets. Note that the red line refers to the error of the ground truth, which is $109.6 $ micron from the precision error of the 3D printer. The model is justified by specifying the error of ground truth which matters while it goes to the 100-micron scale. The mean of L1 loss for training and test set is 0.2381mm and 0.2811mm, respectively. The mean of L2 loss for all training and test sets is 0.0306mm and 0.03208mm. In other words, the DenseTact sensor performs the shape reconstruction with an absolute mean error of 0.28mm. 

\subsection{Evaluation with pose tracking}
\begin{figure}[t!]
      \centering
          \vspace{0.2cm}

      \includegraphics[width = 2.92 in ] {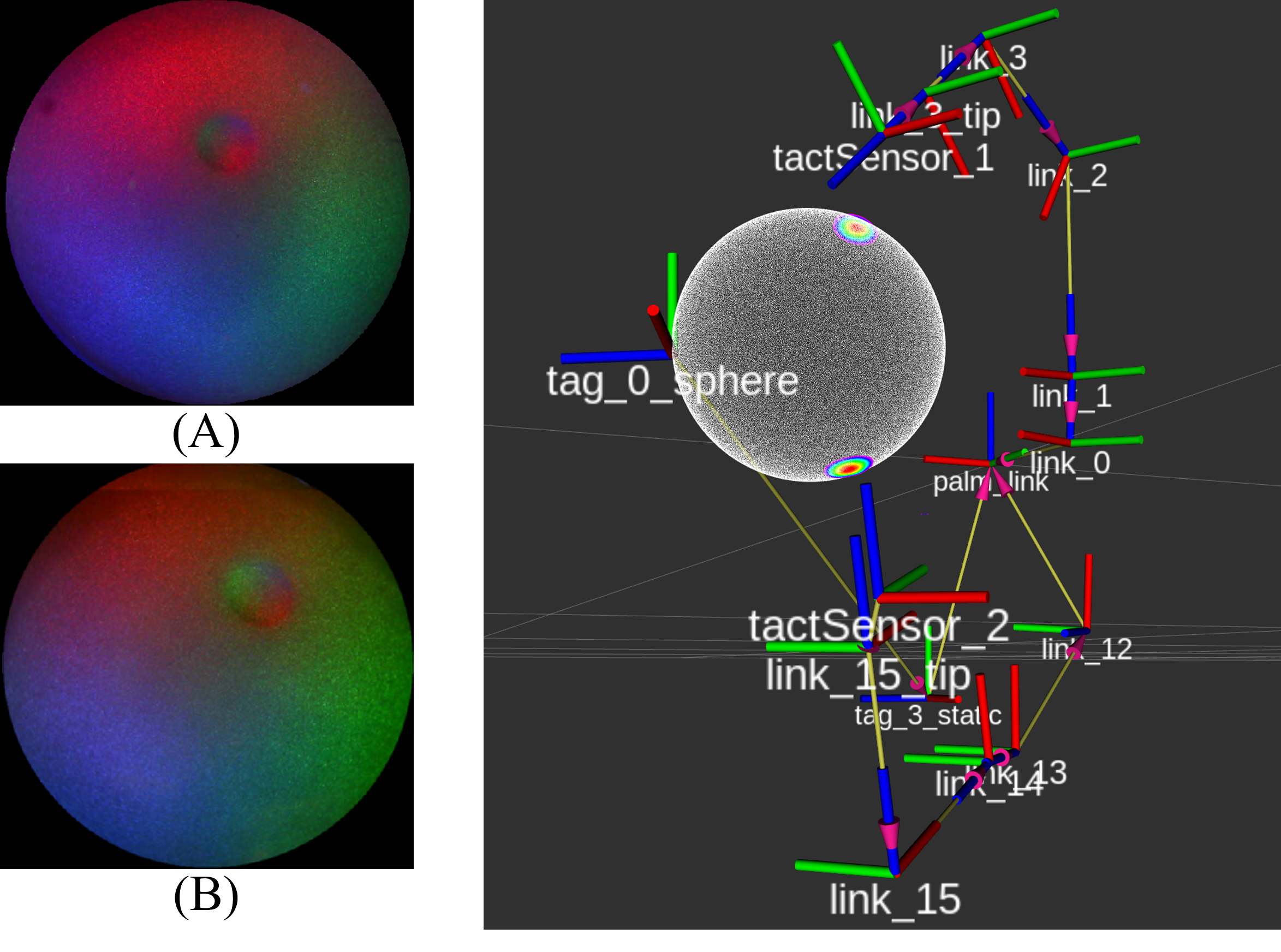}
      \caption{ \textbf{Evaluation with ICP.} Images from two DenseTacts (left) and point clouds of object and DenseTact (right) for the grasp shown in \figref{fig:process_a}.}
      \label{fig:evaluation}
          \vspace{-0.7cm}

  \end{figure}

\edits{
 We evaluate the sensor by tracking the pose of a known object by pinching the object with DenseTact-attached Allegro hand. We compare the 3D reconstructed point cloud from two DenseTact with known spherical object through ICP (Iterative Closest Point) algorithm. Two DenseTacts are installed in the Allegro hand while a fiducial marker (apriltag) is attached to the palm of the hand with a known pose (see \figref{fig:process_a}). To get the reference frame between Densetacts and grasped object, we attached a fiducial on the object and frames are configured through encoder of allegro hand and fiducial configuration from outside camera. The point-to-point ICP registration algorithm in open3D library \cite{Zhou2018} was used to track estimated pose of the object from DenseTact point cloud. \figref{fig:evaluation} shows the point cloud of the object (white points in the right image) and point clouds from DenseTacts (colored point cloud in right image). Image A and B show the input images from upper and lower DenseTact. We evaluated the sensor by measuring first RMS error and fitness between detected DenseTact point cloud and corresponding points from object point cloud. Fitness refers to the ratio of the number of inlier correspondences and the number of DenseTact point cloud. After 23 grasping trial, the average fitness score was 0.597 ($\sigma$ = 0.238) and average RMS error was 0.037184 ($\sigma = 0.00276$). Note that the average RMS error after 200 iterations of ICP was 0.0211. }

\subsection{Discussion}
Fig. \ref{fig:l1result} shows that the average absolute value of radial depth of all test and training sets have small and similar variance. Observed modality on the training set refers to the different types of indicators -- one hole, two holes, and one hole with a pre-built indenter. Future work includes increasing the size and variability of the dataset to reduce bias. 
Possible sources of error for the sensor performance include ground truth error from the 3D printer and human error during dataset collection -- as the indenter was manually rotated in the A and C axis in Fig. \ref{fig:datacollect}. Additional rotation error could possibly come from the error of the stepper motor, however, the error ($0.18 ^\circ$ per step) is relatively small comparing to the human error. The sensor has been shown to be durable enough to use in real applications.
\edits{The sensor performed well without any noticeable change on the sensor after 30,000 pushes and measurements.}



\section{Conclusion}
\label{sec:conclusion}
This work presents a tactile sensor DenseTact, an anthropomorphic hemispherical-shaped sensor capable of reconstruction of the entire sensor surface. \edits{The sensor is shown to be durable, enduring more than 30,000 measurements without noticeable change.}
The sensor is calibrated with high-resolution contact, accounting for relative accuracy and uncertainty in the ground truth.
A neural network is leveraged to model the depth map given the input image and showed that with this off-the-shelf encoder-decoder based network with the pre-trained model is capable of accurate depth reconstruction with \edits{our} training dataset.
The sensor was able to achieve an average of 0.28mm depth difference on the test set. Future work includes leveraging multiple LEDs with different LED configurations to increase the accuracy of the sensor. Different depth maps of single observation from rotations of the LED configuration will increase the accuracy of the sensor. Next sensor design iterations will be more adaptive in terms of size scalability and different sensor shapes that improve versatility. Furthermore, future implementations of the network structure are planned that will be specialized for the DenseTact sensor. In future work, the contact force distribution will be extracted based on the elastomer's deformation along with the reconstructed shape, ensuring the sensor's ability to sense high-resolution forces compared to other sensors.



  \bibliographystyle{./IEEEtran} 
  \bibliography{./IEEEexample}

\end{document}